# AUTOMATED CALIBRATION OF MOBILE CAMERAS FOR 3D RECONSTRUCTION OF MECHANICAL PIPES

Reza Maalek* (reza.maalek@kit.edu)
*Karlsruhe Institute of Technology, Karlsruhe, Germany*

Derek D. Litchi (ddlichti@ucalgary.ca)
*University of Calgary, Calgary, Canada*

\* *Corresponding author*

*Abstract*

*This manuscript provides a new framework for calibration of optical instruments, in particular mobile cameras, using large-scale circular black and white target fields. New methods were introduced for (i) matching targets between images; (ii) adjusting the systematic eccentricity error of target centers; and (iii) iteratively improving the calibration solution through a free-network self-calibrating bundle adjustment. It was observed that the proposed target matching effectively matched circular targets in 270 mobile phone images from a complete calibration laboratory with robustness to Type II errors. The proposed eccentricity adjustment, which requires only camera projective matrices from two views, behaved synonymous to available closed-form solutions, which require several additional object space target information a priori. Finally, specifically for the case of the mobile devices, the calibration parameters obtained using our framework was found superior compared to in-situ calibration for estimating the 3D reconstructed radius of a mechanical pipe (approximately 45% improvement).*

## Introduction: Photogrammetric Calibration

Photogrammetric Calibration of Optical instruments such as mobile cameras is the process of modeling and reducing the effects of instrumental systematic errors in the acquired data. The process involves the estimation of the interior orientation parameters (IOPs) -and in the case of self-calibration the exterior orientation parameters (EOPs)- of the camera, given several point correspondences between two or more image views. Camera calibration, hence, requires:

(1) an efficient procedure to define exact point correspondences between images; and

(2)    an appropriate geometric camera model to describe the IOPs.

To find exact point correspondences, centers of circular targets, such as those shown in the calibration laboratory of Fig. 1, are almost exclusively utilized in high-precision close-range photogrammetry applications (Luhmann, 2014). Circular targets are geometrically approximated by ellipses in images. They offer several advantages such as a unique center, invariance under rotation and translation, and low cost of production. As the number of targets as well as images increase, on one hand, the manual matching of corresponding targets becomes more tedious, time consuming and impractical. On the other hand, automated matching using only the geometric characteristics of ellipses, runs the risk of mismatches (Type II errors). Therefore, coded targets were recommended to reduce the effects of Type II errors during target matching (Shortis and Seager, 2014). Coded targets, however, still require the design of a distinct unique identifier as well as manual labeling of targets individually, which may serve to be impractical in large target fields. A fully automated process to identify and match targets with minimal manual intervention is, hence, desirable for larger target fields with larger number of images.

In addition, even though the circular targets in the object-space are projected as ellipses in the image plane, the centers of the best fit ellipses in the images do not necessarily correspond to the actual center of the circular target. This is commonly referred to as the eccentricity error, which is a consequence of the projective geometry. The systematic eccentricity error suggests that for two corresponding ellipses in two or more image views, the 3D reconstructed centers of the best fit ellipses, even in the absence of random or systematic errors, will not correspond to the center of the original circular targets (Luhmann, 2014). Hence, the systematic eccentricity error in images must be corrected especially in applications, requiring high-precision metrology.

The other important consideration for camera calibration is the selection of the appropriate geometric camera model. Especially for new optical instruments such as newly released mobile phone cameras, the extent of the impact of the additional parameters, including the terms required to correct radial lens distortions, must be evaluated. In this study, the goal is to model the IOPs and the requirements for metric calibration of 4K video recordings, acquired using three of the latest mobile phone cameras, namely, the iPhone 11, the Huawei P30 and the Samsung S10. The real-world impact of the calibration parameters on estimating the geometric parameters of cylinders representing mechanical pipes in support of as-built documentation of construction sites will also be examined.

To this end, this study focuses on: (i) developing an automated and robust method to detect and match circular targets between video images; (ii) providing a simple approach to correct the eccentricity error; (iii) identifying the geometric error modeling requirements of the mentioned mobile cameras; and (iv) determining the extent of the impact of the camera geometric modeling on the accuracy of a 3D reconstructed mechanical pipe.

## LITERATURE REVIEW

The review of previous literature has been divided into two main categories, namely, matching conics between images, and geometric models for mobile camera calibration. The two are further explained in the following.

*Matching Conics between Images*

Given the camera matrices of two views, $\mathbf{P}$ and $\mathbf{P}'$, the necessary and sufficient conditions for matching conics is given by, $\Delta$, as follows (Quan, 1996):

$$\begin{cases} \det(\mathbf{V}) = \det(\mathbf{A} + \lambda\mathbf{B}) = I_1\lambda^4 + I_2\lambda^3 + I_3\lambda^2 + I_4\lambda + I_5 \\ \mathbf{A} = \mathbf{P^T C P} \\ \mathbf{B} = \mathbf{P'^T C' P'} \\ \Delta = I_3^2 - 4I_2I_4 = 0 \end{cases} \tag{1}$$

where $\mathbf{C}$ and $\mathbf{C}'$ are the conic's algebraic matrices corresponding to views $\mathbf{P}$ and $\mathbf{P}'$, $\mathbf{V}$ is the characteristic polynomial of matrices $\mathbf{A}$ and $\mathbf{B}$, $I_j: j = 1 \dots 5$ are the coefficients of the determinant of $\mathbf{V}$, i.e. $\det(\mathbf{V})$, and $(.)^T$ denotes matrix transpose. Equation (1) shows that for two matching conics between two views, $\Delta$ is equal to zero (or very close to zero in the presence of measurement errors). The problem of automated matching of conics between two images, hence, requires the automated: (i) detection of conics; and (ii) the determination of camera matrices. A comprehensive discussion of available and novel methods for detecting non-overlapping ellipses from images was given in (Maalek and Lichti, 2020a). The remainder of this section, hence, focuses on the latter requirement, i.e. automated methods of recovering camera matrices, given only point correspondences.

*Recovering Camera Matrices.* Fundamental matrices are widely implemented in computer vision to provide an algebraic representation of epipolar geometry between two images. Given a sufficient set of matching points (at least seven), the fundamental matrix, $\mathbf{F}$, can be estimated (Hartley and Zisserman, 2000). An important property of fundamental matrices is that the estimated $F$ between two views is invariant to projective transformation of the object space. Therefore, the relative camera matrices can be obtained directly by the fundamental matrix up to a projective ambiguity (Luong and Viéville, 1996). If the camera model is assumed unchanged between two views, the reconstruction is possible up to an affine ambiguity. In case an initial estimate of the IOPs are available, the fundamental matrix can be decomposed into the essential matrix, $\mathbf{E} = \mathbf{K^T F K}$, where $\mathbf{K}$ represents the matrix of intrinsic camera parameters. Given matrix $\mathbf{K}$, the relative orientation between two cameras can be retrieved through singularvalue decomposition of the essential matrix (Hartley and Zisserman, 2000) with only five point correspondences (Stewénius et al., 2006). In such cases, it is possible to recover the reconstruction up to a similarity transform, i.e. an arbitrary scale factor. The two-view process can also be extended to multiple image views. In fact, the camera matrices of $m$ images can be recovered, given at least $m - 1$ pair-wise fundamental matrices and epipoles, using the projective factorization method of (Sturm and Triggs, 1996). The latter is an example of a global reconstruction framework. In practice, however, a sequential reconstruction and registration of new images typically produces more reliable results, due to the flexibilities and control, inherent in the incremental improvement of the reconstruction solution (Schönberger, 2018).

*Sequential Structure-from-Motion (SfM).* Structure from motion is the process of retrieving camera IOPs and EOPs subject to a rigid body motion (i.e. rotation and translation (Ullman, 1979)). The overview of a typical sequential SfM is comprised of the following steps:

(1) Detect and match features between every pair of images and determine overlapping images. The point correspondences are typically obtained automatically using established computer vision feature extraction and matching methods such as the scale-invariant feature transform (SIFT; (Lowe, 2004)), speeded up robust features (SURF; (Bay et al., 2008)), or their variants.

(2) Start with an initial pair of images (typically the two images obtaining the highest score for some geometric selection criterion; see (Schönberger, 2018).

(3) Estimate the relative orientation, and camera matrices between the two images using the corresponding feature points. Note that an initial estimate of the IOPs is typically required for this stage.

(4) Triangulate to determine 3D coordinates of the corresponding points.

(5) Perform bundle adjustment to refine IOPs and EOPs.

(6) For the remaining images, perform the following:

    (a) Add a new overlapping image to the set of previous images.

    (b) Estimate the relative orientation parameters of the new image from the existing overlapping feature points.

    (c) Determine additional matching feature points and triangulate them to estimate the object space coordinates of the new feature points.

    (d) Perform bundle adjustment to refine the IOPs and EOPs.

    (e) Perform steps 6a through 6d until all images are examined.

The output of SfM is a set of EOPs and IOPs, along with a set of image features, which were used for sparse reconstruction. Several approaches as well as software packages exist that perform different variants of SfM. In this study, COLMAP, an open source software, comprised of many computational and scientific improvements to traditional SfM methods, documented in (Schönberger, 2018), is utilized.

*Geometric Models for Mobile Camera Calibration*

Mobile phone cameras can be considered as pinhole cameras for which the collinearity condition can be used to model the straight-line relationship between an observed image point, $(x, y)$, its homolgous object point $(X, Y, Z)$ and the perspective center of the camera, $(X^c, Y^c, Z^c)$ as described in (Luhmann et al., 2013). Random error departures from the hypothesized collinearity condition are modelled as additive, zero-mean noise terms $(\varepsilon_x, \varepsilon_y)$ while $(\Delta x, \Delta y)$ represent systematic error correction terms. The latter comprise the models for radial lens distortion and decentering distortion. Radial lens distortion, which is by far the larger of the two distortions, is most often modelled with three terms of the standard polynomial (e.g. (Luhmann et al., 2016)), though higher-order terms have been demonstrated to be required for wide-angle lenses (Lichti et al., 2020). Images collected with modern mobile phone cameras are generally corrected for radial lens distortion. However, the extent of the correction and the metric impact, if any, on object space reconstruction is not known and a subject of this investigation.

<center>METHODOLOGY</center>

The proposed method for metric calibration of mobile cameras is formulated as follows:

(1) **Calibration data collection:** which involves the method for data collection from the calibration laboratory (Fig. 1).
(2) **Circular target center matching**: which consists of the following three stages:
  (a) Estimation of camera projective matrices (Fig. 2a).
  (b) Automated ellipse detection from images (Fig. 2b).
  (c) Automated ellipse matching between images (Fig. 2c).
  (d) Correction of ellipse eccentricity error (Fig. 2d).
(3) **Self-calibrating bundle adjustment**

Each section is introduced in more detail in the following.

*Calibration Data Collection*

This study focuses on calibration of mobile cameras using video sequences. To improve the precision of the self-calibration and prevent projective compensation (coupling), the recordings must (i) capture depth variation in the target field; (ii) be convergent; and (iii) rotate 90° about the camera's optical axis (landscape and portrait images). One corner of a complete calibration laboratory is utilized, Fig. 1a, which consists of multiple targets attached to two right angled intersecting walls. The two intersecting planar walls are utilized to create depth variation in the target field. A ladder is used to collect convergent images of the scene starting from the bottom of the ladder facing the ceiling (Fig. 1b), ending at the top of the ladder facing the floor (Fig. 1c). At the top, the camera is rotated 90° and the video was recorded in the reverse order (Fig. 1d and Fig. 1e).

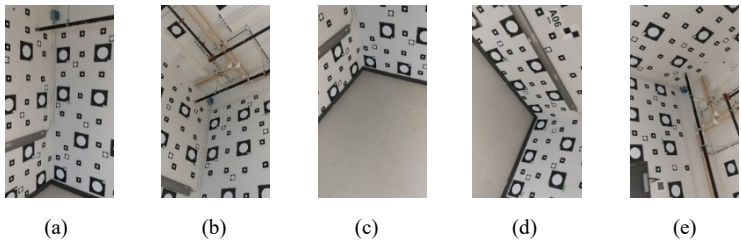

(a)          (b)          (c)          (d)          (e)

FIG. 1. Process to collect video data for calibration: (a) corner of the calibration laboratory; (b) bottom of the ladder facing to the ceiling in portrait; (c) top of the ladder facing the floor in portrait; (d) top of the ladder facing the floor in landscape; and (e) bottom of the ladder facing the ceiling in landscape.

*Circular Target Center Matching*

The problem of circular target matching for calibration requires: (i) estimation of camera projective matrices (Fig. 1f); (ii) automated detection of ellipses from each image (Fig. 2a); (iii) correct matching of corresponding ellipses between different images (Fig. 2b); and (iv) correction of the eccentricity error of the ellipses' centers (Fig. 2c). These steps are discussed in more detail in the following.

*Estimating the Camera Projective Matrices.* Based on the discussions provided in Section 2.1, an initial estimate of the camera projective matrices (comprised of IOPs and EOPs) can be obtained using an SfM framework, such as COLMAP. Here, **Algorithm 1:**

**Refining Camera Matrices**, is proposed to further improve the estimated camera matrices, retrieved from the outputs of COLMAP's sparse reconstruction:

(1) For each image, determine the features used for sparse reconstruction from COLMAP.

(2) Between every two images, $i \in 1 \ldots n-1$ and $j \in i+1 \ldots n$, where $n$ is the total number of images, with overlapping features, perform the robust least median of squares (LMedS; (Rousseeuw and Leroy, 1987)) fundamental matrix estimation with the subsample bucketing strategy of (Zhang, 1998) and retrieve the inlier features and fundamental matrix.

(3) For the inlier matching feature points of step 2, perform the robust triangulation using LMedS with random subsampling. Identify the inlier feature points for every given sparsely reconstructed point.

(4) Perform bundle adjustment on only the inlier features of steps 2 and 3 (i.e. weight all outlying features as zero) to retrieve the refined EOPs and IOPs.

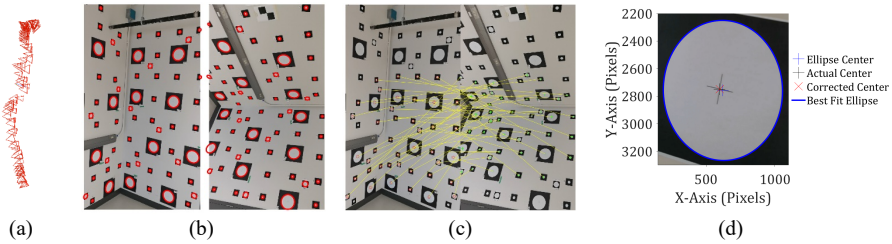

(a)  (b)  (c)  (d)

FIG. 2. Outputs of the proposed steps to acquire matching target centers between images: (a) estimated camera position and orientation (EOPs) using Algorithm 1; (b) detected ellipses using the method of (Maalek and Lichti, 2020b) for two images; (c) automatic matching of corresponding target centers using Algorithm 2; and (d) correction of the estimated target center in image plane using Algorithm 3.

As a point of reference, Fig. 2a illustrates the estimated camera positions and orientations of a sequence of images collected with the proposed strategy using Algorithm 1. Fig. 3 shows the refinement of the matched features using Algorithm 1. As observed, two mismatched features, represented by red and green ovals were correctly removed. The red oval was a feature point that did not satisfy step 2 of Algorithm 1, whereas the mismatched feature of the green oval was removed using step 3. In this example, even though the refinement is marginal, two mismatches out of around 300 correct matches, it can still negatively affect the results of the estimated IOPs and EOPs especially when most images contain mismatches.

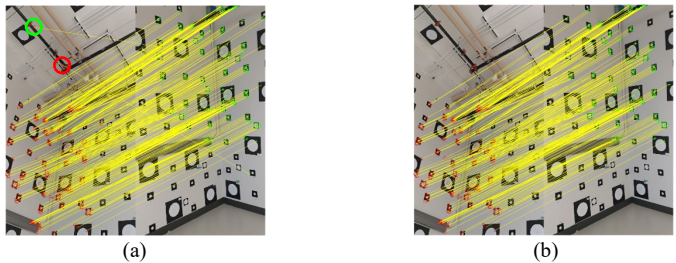

(a)  (b)

FIG. 3. matched features between images: (a) results from COLMAP; (b) results of refinements using Algorithm 1.

*Automated Ellipse Detection from Images.* The robust ellipse detection method presented in (Maalek and Lichti, 2020a) is used to detect non-overlapping ellipses of the projected circular targets. The method was shown to provide ellipse detection with superior robustness to both Type I and Type II errors, compared to the established ellipse detection methods of (Fornaciari et al., 2014) and (Pătrăucean et al., 2017). Once the ellipses are detected, the best fit geometric parameter vector of each ellipse is estimated using the new confocal hyperbola ellipse fitting method, presented in (Maalek and Lichti, 2020b). The geometric parameters of the ellipse, i.e. center, semi-major length, semi-minor length and rotation angle, are then converted to algebraic form, which are then transformed into matrix form to be used for conic matching through equation (1).

*Automated Ellipse Matching between Images.* Equation (1) is utilized here to verify possible matching conics between two images. A brute force matching strategy suggests checking the correspondence condition, $\Delta$ of equation (1) for every detected ellipse of an image to all ellipses of all other images, which can become computationally expensive in larger number of images with many targets. In addition, the $\Delta$ calculated using equation (1) is not necessarily equal to zero in the presence of systematic and random measurement errors (Quan, 1996). Therefore, a threshold is required on $|\Delta|$ to model the matching uncertainties in the presence of measurement errors. An arbitrarily selected threshold for the correspondence condition, $\Delta$, will, however, almost guarantee either mismatches (Type II errors) or no matches (Type I errors).

Here, instead of comparing all conics together in a brute force fashion, first, only a select set of candidates are considered, whose centers in both images satisfy an adaptive closeness constraint on the corresponding epipolar distance. Amongst the available ellipses satisfying the epipolar constraint, that achieving the smallest $|\Delta|$ is chosen as the matching conic. This process is attractive for two reasons. Firstly, since the robust fundamental matrix between two images is already computed automatically using the LMedS method in Algorithm 1 (step 2), the agreement of the inlier points to the estimated fundamental matrix can be used as a basis to compute the threshold for the epipolar constraint on the centers as well (no pre-defined subjective threshold required). Secondly, the ellipse matching using the correspondence condition of equation (1) requires no threshold. To formulate the proposed process, **Algorithm 2: Automated Matching of Ellipses**, is provided as follows:

(1) Estimate the camera projective matrices using Algorithm 1 (or any other preferred method).
(2) Detect the ellipses using the robust non-overlapping ellipse detection of (Maalek and Lichti, 2020a).
(3) For each detected ellipse, find the best fit ellipse using the method of (Maalek and Lichti, 2020b), and construct the equivalent algebraic conic matrix.
(4) For every pair of overlapping images, $i \in 1 \dots n-1$ and $j \in i+1 \dots n$, where $n$ is the total number of images, find the matching ellipses as follows:
   (a) From the LMedS algorithm of step 1:
      (1) Use the solutions obtained for the fundamental matrix, $\mathbf{F_{ij}}$.
      (2) Calculate the standard deviation of the epipolar distance (Hartley and Zisserman, 2000), $\sigma_{ij}$, of the inlier matches.
   (b) For each ellipse center in image $i$, $c_{ik}$, find the ellipse centers in image $j$, $c_{jl}$, that satisfies the following condition.

$$D\left(c_{ik}, c_{jl}\right) = \sqrt{d\left(c_{jl}, F_{ij}c_{ik}\right)^2 + d\left(c_{ik}, F_{ij}^T c_{jl}\right)^2} \leq \sigma_{ij}\sqrt{\chi_{0.975,2}^2} \qquad (2)$$

where $D\left(c_{ik}, c_{jl}\right)$ is the epipolar distance between image points $\left(c_{ik}, c_{jl}\right)$, and $\chi_{0.975,2}^2$ is the chi-squared cumulative probability distribution function with probability 97.5% and degrees of freedom 2 for 2D data.

  (c)   For all ellipse candidates satisfying equation (2) between two images, perform the following:

  (1)   Select all related matching ellipses between the two images. For instance, if ellipse labels 5 and 6 of image $j$ satisfy equation (2) for ellipse labeled 4 of image $i$, all other ellipses of image $i$ that satisfy equation (2) for ellipses labeled 5 and 6 of image $j$ should also be selected and so on.

  (2)   Calculate $\Delta$ from equation (1) between the ellipse candidates (from the previous step) of images $i$ and $j$.

  (3)   Two ellipses are considered matching if and only if both ellipses achieved the minimum $|\Delta|$ for each other. For instance, if the ellipse labeled 5 of image $j$ achieved the smallest $|\Delta|$ for label 4 of image $i$, the two will only be matched if the ellipse labeled 4 of image $i$ achieves the smallest $|\Delta|$ for label 5 of image $j$.

  (5)   Using the matched ellipses of step 4 in two-views, determine the matching ellipse between all other images.

  (6)   Using the multi-view correspondences identified in step 5, for each ellipse, perform the following steps to reduce the impact of Type II errors (mismatches):

  (a)   Use the camera projective matrices estimated in step 1.

  (b)   Perform robust multi-view triangulation on the ellipse centers using LMedS (or any other preferred robust method).

  (c)   Retain only the inlier set of corresponding ellipse centers.

  (7)   The inlier ellipse correspondences are the final set of matched ellipses.

Step 4b of Algorithm 2 performs the pairwise conic matching condition on only a selected set of inlier ellipse centers that lie near the epipolar line. Furthermore, instead of choosing an arbitrary threshold for $\Delta$, the minimum of $|\Delta|$ is used. To further reduce the effects of Type II errors (mismatching targets), the robust triangulation using LMedS is performed.

*Correction of Ellipse Eccentricity Error.* Modeling the eccentricity error for circular targets has been the subject of investigation, especially in high-precision metrology (Ahn et al., 1999; He et al., 2013; Luhmann, 2014). Closed formulations of the eccentricity error in both the image plane and object space exist (Ahn et al., 1999; Dold, 1996). The available correction formulations for the errors in the image plane, however, require the knowledge of the target's object space parameters, such as the radius of the target, the object space coordinates of the center and the normal vector of the circular target's plane. This information, however, cannot be trivially retrieved. Even if a reliable external measurement of the object space exists, the additional constraint will be undesirable in free-network self-calibration practices, which is the focus of this study. In the following, a process is presented to retrieve the projection of the true center of the circular target onto the image, given only the matching target parameters in two-views.

Following the formulation of equation (1), Quan (Quan, 1996) proposed a process to acquire the equation of the object space plane where the conic lies, given two camera projection matrices and the conics' algebraic matrices in the image plane. The subsequent object space conic's equation for the matching conics can then be retrieved by intersecting the plane with the cone's equation (see equation (1)). Ideally, the object space conic should be represented by a circle in the case of circular targets; however, due to measurement errors and uncertainties in the estimation of the projection matrices, the object space conic might be an ellipse. The object space center of the ellipse (or circle) can be directly extracted from the retrieved conic's equation. The corrected center in the image can, hence, be extracted by back projecting the object space center of the retrieved ellipse onto the image planes. Since multiple views may observe a given target, the final consideration is to determine the best view for a given image view. To this end, two criteria are used: (i) the convergence angle between the two views; and (ii) the uncertainty of the object space center estimation. Firstly, for a given view, only the image views with an average convergence angle of more than 20° (Schönberger, 2018) are considered. Secondly, for a specific matched target, the uncertainty of the object space center estimation is characterized, here, as the covariance of the 3D reconstructed ellipse centers between a given view and another acceptable view. The covariance is estimated using the formulation provided by (Beder and Steffen, 2006). The two views achieving the minimum determinant of the covariance matrix (MCD) -representing the pair with the minimum uncertainty- are selected as the ideal candidate (Rousseeuw and Leroy, 1987). Intuitively, the latter selection criterion provides the pair of images, whose object space coordinates in the vicinity of the true center are the least uncertain. The process is formulated using **Algorithm 3: Correcting Eccentricity Error**, for each identified target with two or more image view correspondences, as follows:

(1) Retain the triangulated object space coordinate of the target's center from step 6 of Algorithm 2.

(2) Identify all images corresponding to the considered target, say images $1 \dots m$, where $m$ is the number of views corresponding to a given target.

(3) For each image $i \in 1 \dots m$, find the best image pair in $j \in 1 \dots m, i \neq j$, (image pair with the least uncertainty in the estimated target's center) as follows:

    (a) Find all images in $j \in 1 \dots m, i \neq j$ whose average convergence angle from image $i$ is more than 20° (Schönberger, 2018).

    (b) Estimate the covariance matrix of the 3D reconstructed ellipse centers of the views obtained by the previous step using (Beder and Steffen, 2006).

    (c) For image $i$, find the corresponding pair whose determinant of the covariance matrix is minimum (i.e. MCD).

    (d) Repeat the steps 3a to 3c to find a best image pair for all images $1 \dots m$.

(4) For image $k$ with selected best image $l \ \{k \neq l \mid k, l \in 1 \dots m\}$, perform the following to retrieve the corrected ellipse center in image $k$:

    (a) Determine the object space plane parameters, $\mathbf{T}_{kl} = (t_1, t_2, t_3, t_4)^T$, of the conic in space using the method of (Quan, 1996).

    (b) Intersect plane $\mathbf{T}_{kl}$ with the cone $\mathbf{A_k}$ (for image $k$) to find the conic equation as follows ($\mathbf{A}$ explained in equation (1)):

        (1) Parametrically derive the $Z$ coordinate as a function of $X$ and $Y$ using the planes' equation as follows:

$$Z = -\frac{t_1 X + t_2 Y + t_4}{t_3} \tag{3}$$

(2) Substitute the $Z$ into the cone's equation to derive the object space conic matrix of the ellipse, $\mathbf{C_k}$, as follows:

$$[X \quad Y \quad Z \quad 1]\, \mathbf{A_k} \begin{bmatrix} X \\ Y \\ Z \\ 1 \end{bmatrix} \xrightarrow{\text{Substitute } Z} [X \quad Y \quad 1]\, \mathbf{C_k} \begin{bmatrix} X \\ Y \\ 1 \end{bmatrix} = 0 \tag{4}$$

(3) Calculate the geometric center $(X_c, Y_c)$ of the ellipse corresponding to the conic matrix $\mathbf{C_k}$ as follows:

$$X_c = \frac{\det \mathbf{C_{k31}}}{\det \mathbf{C_{k33}}} \quad , \quad Y_c = -\frac{\det \mathbf{C_{k32}}}{\det \mathbf{C_{k33}}} \tag{5}$$

where $\mathbf{C_{k_{ij}}}$ is the 2×2 matrix constructed after removing row $i$ and column $j$ from $\mathbf{C_k}$.

(4) Substitute $(X_c, Y_c)$ into equation (3) to calculate the $Z$ component of the object space 3D coordinates of the center of the ellipse $\mathbf{X_k}=(X_c, Y_c, Z_c)$.

(5) Project the center of the conic back onto the image plane to find the corrected center, $\mathbf{x_k} = \mathbf{P_k X_k}$, where $\mathbf{P_k}$ is the camera projective matrix for image $k$.

(6) Convert the estimated centers into homogeneous coordinates for use in the bundle adjustment.

(5) Once the target's center in all images has been corrected, perform triangulation to correct the object space coordinates of the center.

The results of Algorithm 3 are fed to a free-network self-calibrating bundle adjustment that estimates the EOPs, IOPs and object space coordinates of the targets. Once the EOPs and IOPs are determined, the centers can again be adjusted and recursively fed to the bundle adjustment to refine the results up to the required/satisfactory precision. Furthermore, Algorithm 3 can be utilized before the robust triangulation step of Algorithm 2 (after step 4) so that possible correct matches are not incorrectly rejected (improving Type I errors in target matching). Fig. 4 illustrates the results of using Algorithm 3 within Algorithm 2. In this example, one additional target was correctly matched (amongst a total of 23 overlapping targets) when applying Algorithm 3 within Algorithm 2.

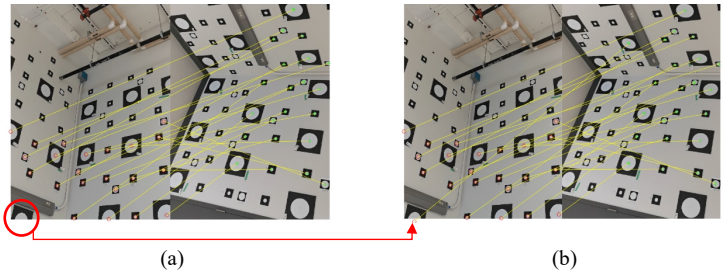

(a)                                              (b)

FIG. 4. Impact of center correction on the results of the target matching between two sample images: (a) without center correction; and (b) with center correction

*Self-Calibrating Bundle Adjustment*

The final step of the algorithm is the self-calibrating bundle adjustment, which is the accepted standard methodology for obtaining the highest accuracy (Luhmann et al., 2016). Provided that the aforementioned design measures have been incorporated into the imaging network, the IOPs will be successfully de-correlated from the EOPs and, thus, recovered accurately. Any outlier observations have been successfully removed by this point of the algorithm, so a least-squares solution can be utilized for the parameter estimation. The IOPs, which are considered network-invariant (i.e. one set per camera), EOPs and object points are estimated. The singularity of the least-squares normal-equations matrix caused by the datum defect is removed by adding the inner constraints (the free-network adjustment; (Luhmann et al., 2013)) since this yields optimal object point precision.

Following the self-calibration adjustment, the solution quality can be examined with several computed quantities. Most important among these are the estimated IOPs and their precision estimates, together with derived correlation coefficient matrices that quantify the success of the parameter de-correlation. The residuals are crucial for graphically and statistically assessing the effectiveness of the lens distortion modelling. The presence of un-modelled radial lens distortion, for example, can be readily identified in a plot of the radial component of the image point residuals as a function of radial distance from the principal point. Moreover, reconstruction accuracy in object space can be quantified by comparing the photogrammetrically-determined coordinates of (or derived distances between) targets with reference values from an independent measurement source.

*Summary of Methods*

The proposed camera calibration framework can be summarized as follows:

(1)  Determine an initial estimate of the EOPs and IOPs using available SfM methods or software packages (here, COLMAP was used).
(2)  Refine the estimated EOPs and IOPs to calculate the modified camera projective matrices for each image view, using Algorithm 1.
(3)  Determine the ellipses in each image using the robust non-overlapping ellipse detection of (Maalek and Lichti, 2020a).
(4)  Match overlapping ellipses between all views using Algorithm 2.
(5)  Adjust the eccentricity error of the estimated ellipse centers of all matched ellipses, using Algorithm 3.
(6)  Perform the proposed free-network self-calibrating bundle adjustment to estimate the EOPs and IOPs.
(7)  Perform steps 4 through 6 with the new EOPs and IOPs until the sets of matched ellipses between two consecutive iterations remain unchanged.

The final set of IOPs is the solution to the calibration.

*Selection of Terms for Radial Lens Distortion*

The accuracy of 3D reconstruction from a camera system affected by lens distortions can be significantly impacted by the choice of systematic error correction terms included in the augmented collinearity condition. The aim is to find a trade-off between goodness-of-fit and allowable bias. One must avoid an under-parameterized model with an insufficient number of terms to describe the distortion profile that can lead to the propagation of bias into other model parameters and optimistic parameter precision. On the other hand, adding more terms than necessary can introduce correlations among model variables that can inflate the condition number of the normal-equations matrix and, in turn, degrade reconstruction accuracy. To this end, a process similar to that described in (Lichti et al., 2020) was employed. Using the final set of matched targets, an initial self-calibration solution was performed without any lens distortion parameters. The interior geometry of the camera in this adjustment is described only by the principal point and principal distance. Graphical analyses of the estimated residuals, supported with statistical testing and information criteria, are utilized to make an informed decision about the coefficients to be added. In particular, the radial component of the image point residuals, $v_r$, plotted as a function of radial distance from the principal point, $r$, is graphically assessed. A single parameter is then added to the model and the self-calibration is recomputed. The process is repeated until no systematic trends remain.

*Method of Validation of Results*

The effectiveness of Algorithm 2 requires the quantification of the quality of ellipse matching between different images. Here, the four main metrics, commonly used to determine the quality of object extraction algorithms, namely, precision, recall, accuracy, and F-measure (Olson and Delen, 2008), are utilized. To measure the accuracy of estimated parameters, such as center adjustment in Algorithm 3, the Euclidian distance (or L2-norm) of the estimated parameters from the final ground truth parameters were used. The ground truth in each experiment was determined manually.

<center>EXPERIMENT DESIGN</center>

Four experiments are designed to assess the effectiveness of the proposed methods used in this study, namely, evaluation of ellipse eccentricity adjustment, quality assessment of ellipse matching, and evaluation of impact of calibration on pipe reconstruction. The four experiments are explained in more detail in the following.

*Evaluation of Ellipse Eccentricity Adjustment*

Algorithm 3 was developed to correct the eccentricity error of the estimated center of the targets in images due to projective transformation. This experiment is designed to evaluate the effectiveness of the proposed method in practical settings. To this end, 28 images were recorded from a single target, using a calibrated Huawei P30. The EOPs of each image view were estimated using COLMAP and shown in Fig. 5a. The ground truth target center in each image was manually determined. The scale of the 3D reconstruction was manually defined using the radius of the circular target. The precision of the 3D reconstructed center using the ground truth image centers and estimated EOPs was 0.13mm. The ground truth center, the estimated best fit ellipse center, and the adjusted center using Algorithm 3 for one sample image are shown in Fig. 5b. Two experiments were designed to: (i) quantify the impact of center adjustment on the accuracy of the 3D reconstructed center; and (ii) evaluate its performance compared to the closed formulation of eccentricity error by (Luhmann, 2014).

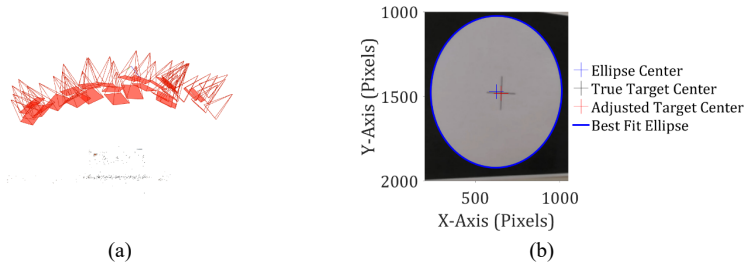

(a)                                                           (b)

FIG. 5. Design of the evaluation of ellipse eccentricity adjustment experiment: (a) EOPs for the 28 image views; and b) ellipse center, adjusted target center, and true target center for a sample image and target.

*Impact of Center Adjustment on the 3D Reconstruction.* In this experiment, the impact of adjusting the eccentricity error compared to the estimated best fit ellipse center (unadjusted) on the accuracy of the 3D reconstructed center of the target is evaluated as the number of camera views increase from 2 to 28. Since different combination of views will generate different reconstruction results, for a given number of (say $k$) image views, $N$ different combinations of $k = 2 \dots 28$ images are selected. To this end, for $k$ image views, the following steps are carried out:

(1) Randomly select $N$ different combination of $k$ images from the 28 images.
(2) For each set of $k$ images, using the estimated best fit ellipse center (no adjustment), adjusted center using Algorithm 3, and camera projection matrices (see Fig. 5a), perform triangulation (Hartley and Zisserman, 2000) and determine the object space position of the target centers.
(3) Calculate the Euclidian distance between the object space coordinates of the adjusted and unadjusted centers (separately) from the ground truth center.
(4) For the given number of image views, $k$, record the mean of the $N$ distances obtained from step 3 for the adjusted and unadjusted centers.

Here, $N = 50$ combinations are selected.

*Comparison of Eccentricity Error between Algorithm 3 and Luhmann's Formula.* A closed formulation of the center eccentricity error in the image plane was provided in (Luhmann, 2014), given the EOPs and IOPs of the view, 3D object space target center, object space target radius, and target plane's normal. To determine the eccentricity error using Luhmann's formula, the ground truth 3D object space center, radius as well as the plane normal were used. For each image, the eccentricity errors (in pixels) were calculated using our method (Algorithm 3), and Luhmann's formula. For each image, both eccentricity errors in each image were then compared to the ground truth eccentricity error and reported.

### Quality Assessment of Ellipse Matching

This experiment is designed to assess the quality of the ellipse matching results, obtained by Algorithm 2. To this end, 1.5 minutes of 4K video is recorded as per the presented data collection method, using the Samsung S10, the iPhone 11, and the Huawei P30. 90 images are extracted from each recording at 1 frame per second (fps). The initial EOPs, and IOPs were extracted from COLMAP and refined using Algorithm 1. The ellipses of the 270 images were extracted using the method of (Maalek and Lichti, 2020a). The camera projective matrices as well as the detected ellipses for each image were then fed to Algorithm 2 to determine the matching ellipses between different images. The quality of matching was quantified for three settings, namely, Algorithm 2 without robust triangulation, Algorithm 2 with robust triangulation and Algorithm 2 with robust triangulation and corrected centers.

*Evaluation of Impact of Calibration on Pipe Reconstruction*

The broader objective of this study pertains to the application of mobile phone cameras for 3D reconstruction of pipes. To this end, mechanical mock pipes were professionally installed at one corner of the calibration laboratory, shown in Fig. 6. This experiment was designed to assess the effectiveness of the proposed calibration process in estimating the radius of the pipe of interest after 3D reconstruction. Here, the accuracy of the estimated radius using our pre-calibration process and lens distortion modeling was compared with that obtained using COLMAP's default SfM process. COLMAP's default SfM process involves an in-situ automatic calibration, comprised of the first two terms of the radial lens distortion parameters, using the exchangeable image file (Exif) data as the initial IOP estimation. This latter process from here on is referred to COLMAP-radial.

A 60 second 4K video is recorded around one pipe using each of the Huawei P30, iPhone 11, and the Samsung S10. The recording was divided into two, 30 second videos (at 1fps) in portrait and landscape modes. The camera was rotated 90° about its optical axis so as to provide COLMAP-radial a fair opportunity to calibrate the instruments without possible projective coupling. A dense 3D reconstruction is then carried out using COLMAP once with the IOPs obtained by COLMAP-radial and then again with our target-based calibration framework.

The final consideration for the 3D reconstruction is to define the scale. Since the accuracy of estimating the radius of the cylinder is being considered, it is important to define the scale of the 3D reconstruction independent from the cylinder's radius. The scale of the reconstruction is, hence, defined using the distance of two of the targets behind the mock pipes (see Fig. 6b). The ground truth distance between the targets was determined using the HDS 6100 terrestrial laser scanner (TLS). The following process was then performed to determine the scale:

(1) Detect the ellipses in images using ellipse detection of (Maalek and Lichti, 2020a).
(2) Match the detected ellipses between images using Algorithm 2.
(3) Select two of the targets with the maximum number of image views.
(4) Adjust the center eccentricity error of the two targets in each image view using Algorithm 3.
(5) Triangulate to determine the 3D coordinates of the two targets.
(6) Determine the ground truth 3D coordinates of the center of the two targets in the TLS point cloud using the method presented in (Lichti et al., 2019).

(7)   The scale is defined by the ratio between the ground truth distance (step 6) and the 3D reconstructed distance (step 5) of the two targets.

The radius of the cylinder of interest for the scaled 3D reconstruction (Fig. 6c) as well as the TLS point cloud (Fig. 6d) were then calculated using the robust cylinder fitting of (Maalek et al., 2019). The accuracy of the estimated radius is reported for all seven cases (two cases per camera and one case for TLS).

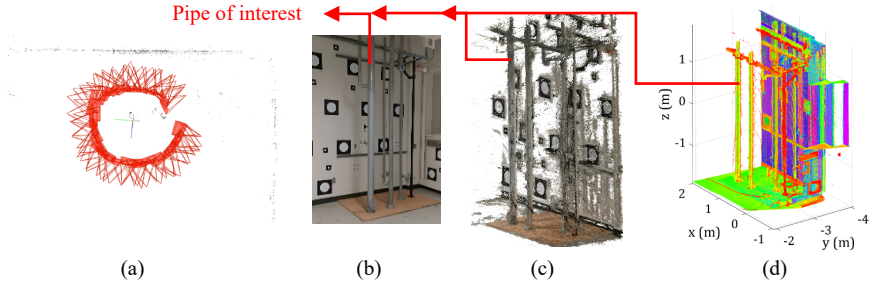

(a)                    (b)                    (c)                    (d)

FIG. 6. Design of the "evaluation of impact of calibration on pipe reconstruction" experiment: (a) sample path around the pipe of desire; (b) sample image of the pipe with the targets on the background wall. 3D Point cloud of the mock pipes using: (c) SfM 3D reconstruction; and (d) HDS6100 TLS.

The summary of the four experiments are provided in Table I.

TABLE I. Summary of the designed experiments.

| Experiment Description | | Type of data | Purpose |
|---|---|---|---|
| Evaluation of ellipse eccentricity adjustment | Impact of center adjustment on the 3D reconstruction | 4K video images using calibrated Huawei P30 | Comparing accuracy of 3D reconstructed center of targets from the best fit ellipse center and the adjusted center of Algorithm 3 |
| | Comparison of eccentricity error between Algorithm 3 and Luhmann's formula | 4K video images using calibrated Huawei P30 | Comparison of eccentricity error using Algorithm 3 and Luhmann's (Luhmann, 2014) closed formula |
| Quality assessment of ellipse matching | | 4K video images using Samsung S10, iPhone 11 and Huawei P30 | Quantifying the quality of matching ellipses between different images |
| Evaluation of impact of calibration on pipe reconstruction | | 4K video images using Samsung S10, iPhone 11 and Huawei P30 with both auto and fixed focus | Assessing the impact of the determined IOPs on the accuracy of estimating the radius of a cylindrical pipe |

EXPERIMENT RESULTS

*Evaluation of Ellipse Eccentricity Adjustment*

*Impact of Center Adjustment on the 3D Reconstruction*. The accuracy of the 3D object space coordinates of the center of the target (shown in Fig. 5) using the best fit ellipse center and the adjusted center using Algorithm 3 was determined as the number of image views increase. Fig. 7 shows the mean accuracy of the estimated centers (for the 50 selected combinations) with no adjustment, and with adjustment. For both adjusted and unadjusted, it can be visually observed that the results of the mean accuracy of the 3D reconstructed center, remains almost constant as the number of image views increase from 5 to 28. The accuracy of the object space 3D coordinates of the center using the adjusted is, however, significantly better than that using the unadjusted ellipse centers. Using all 28 image views, the accuracy of the object space coordinates was 0.64mm and 9.46mm for the adjusted and unadjusted centers, respectively. On average, the results of the unadjusted was approximately 15 times that obtained using the adjusted centers. The result of this experiment demonstrates that reaching sub-millimeter accuracy for the 3D object space coordinates of large target centers becomes possible using the method proposed in Algorithm 3 even for larger targets.

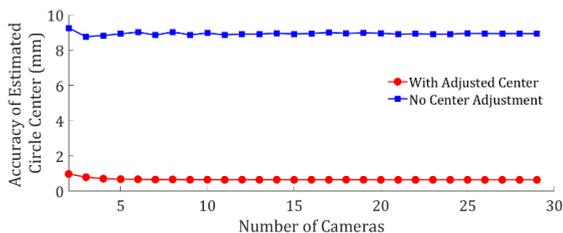

FIG. 7. Impact of adjusting the eccentricity error of elliptic targets in images on the accuracy of the 3D reconstructed center.

*Comparison of Eccentricity Error between Algorithm 3 and Luhmann's Formula*. The eccentricity error, the error between the estimated ellipse center and the actual target center in the image plane, was calculated for each image, using Algorithm 3 as well as Luhmann's closed formulation, given the object space target information. The absolute deviation of the estimated eccentricity (here, referred to as relative eccentricity) using Algorithm 3 and Luhmann's formula from the ground truth eccentricity was calculated. Fig. 8 shows the result of the difference between the ground truth and estimated eccentricity errors for both ours and Luhmann's methods. As illustrated, the results are comparable, however, Luhmann's method achieved slightly better results for the best 22 images. Our method, on the other hand, achieved better results for the worst six images. The average of the relative eccentricity error for all images was 2.84 and 2.41 pixels using Luhmann's, and our method, respectively, which are comparable, but our method provided around 20% improvement compared to Luhmann's. This result is attractive since our formulation requires no a priori knowledge of the target's object space information, which is a requirement for other available eccentricity formulations.

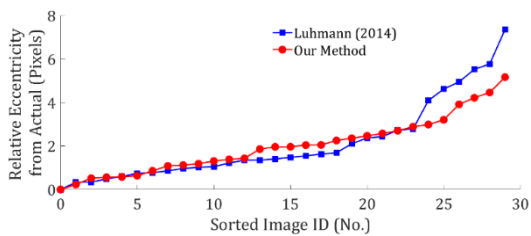

FIG. 8. Eccentricity errors from the ground truth, calculated using Algorithm 3 and (Luhmann, 2014).

*Quality Assessment of Ellipse Matching*

The precision, recall, accuracy and F-measure for the ellipse matching in three settings of (i) Algorithm 2 without robust triangulation (before step 6); (ii) Algorithm 2 complete (with robust triangulation); and (iii) Algorithm 2 combined with Algorithm 3 to also adjust target centers. The quality of the ellipse matching method for the 270 images combined is presented in Table II. The ground truth matches between every two images is manually identified. As observed, Algorithm 2 with robust triangulation achieved a considerably better recall, compared to Algorithm 2 without robust triangulation, which demonstrates its relative robustness to mismatching ellipses (i.e. no Type II errors). The robust triangulation, however, achieved a relatively lower precision compared to when no robust triangulation is performed. This shows that some correct matches are reduced, contributing to an increase in Type I errors. When Algorithm 3 is combined with Algorithm 2 (before the robust triangulation step) to correct the target's eccentricity error, the robustness to Type II was maintained (recall of 100%), and a portion of the correct matches that were eliminated were also recovered (increase of about 3% in the precision). F-measure, which provides a single value to explain both the contributions of Type I and Type II errors, shows that using Algorithm 2 in combination to Algorithm 3 provides the best result. The use of robust triangulation was also found to be necessary to eliminate Type II errors and improve the F-measure, compared to Algorithm 2 without robust triangulation.

TABLE II. Summary of precision, recall, accuracy, and F-measure for the ellipse matching in three variations of Algorithm 2.

| Variations of Algorithm 2 | Precision | Recall | Accuracy | F-Measure |
|---|---|---|---|---|
| Without robust triangulation | **98.44%** | 91.25% | 93.28% | 94.71% |
| With robust triangulation | 93.71% | **100.00%** | 96.15% | 96.75% |
| With robust triangulation and adjusted centers | 96.85% | **100.00%** | **98.08%** | **98.40%** |

*Evaluation of Impact of Calibration on Pipe Reconstruction*

The impact of the pre-calibration using the model and IOPs extracted using our method, compared to COLMAP-radial, for estimating the radius of a mechanical pipe was evaluated. Table III shows the results obtained by COLMAP-radial as well as our pre-calibration for the three cameras. From the results presented in Table III, four observations were made. First, more inlier cylinder points were observed using our pre-calibration, compared to COLMAP-radial for all camera devices (about 1.3 times on average). This suggests that better feature matching was obtained from the same set of images when the correct radial lens distortion parameters and IOPs were used. This is attributed to the fact that the EOPs (particularly fundamental matrices) are impacted by radial lens distortion. In fact, given the same number of point correspondences, known radial lens distortion provides a better estimate of the fundamental matrix, even when the correct radial lens distortion model is considered (see Fig. 3 of (Barreto and Daniilidis, 2005)). The second observation was that the RMSE of the best fit cylinder was better using the pre-calibrated model compared to COLMAP-radial in all three devices, even though the number of inlier observations was higher in the pre-calibrated setting. The average difference was, however, only 0.1mm, which may be considered negligible in many practical applications. Thirdly, the accuracy of the estimated radius was better for all devices using our pre-calibration compared to COLMAP-radial (around 45% better accuracy on average). This demonstrates that our pre-calibration procedure for each device provides a better cylinder reconstruction, compared to the in-situ calibration.

The last observation was that the accuracy obtained using the iPhone 11 was better than that using Huawei P30, which were both better than the Samsung S10. This is most likely attributed to the fact that the average number of detected features per image on the iPhone 11 were higher than that of the Huawei P30, which were both higher than that of the Samsung S10. The higher number of matched features are also observed from the number of inlier cylinder points reported using the iPhone, Huawei, and Samsung, shown in Table III (regardless of the calibration procedure). In fact, more inlier points suggest the existence of more correct point correspondences between different images, which consequentially provides a better estimation of the EOPs, especially with the correct camera model (see explanations given in the literature review around essential matrix).

TABLE III. Evaluation of the impact of the proposed calibration process on estimating the radius of mechanical pipe.

| Device | Type | Accuracy of Radius Estimation (mm) | RMSE (mm) | Number of Inlier Cylinder Points (No.) |
|---|---|---|---|---|
| Huawei P30 | Pre-calibration | 1.2 | 1.2 | 187,225 |
| | COLMAP-radial | 3.1 | 1.3 | 126,694 |
| iPhone 11 | Pre-calibration | 0.5 | 1.0 | 219,297 |
| | COLMAP-radial | 1.4 | 1.1 | 166,238 |
| Samsung S10 | Pre-calibration | 4.1 | 2.3 | 84,122 |
| | COLMAP-radial | 6.5 | 3.4 | 72,542 |

CONCLUSIONS

This manuscript provided a collection of new methods for the automatic calibration of optical instruments in particular mobile cameras. To this end, 4K videos, decomposed into images at 1 fps, were recorded from the calibration laboratory with a redundant set of black and white circular targets of different sizes. The method then utilizes the potential of SfM for the sequential calibration and reconstruction of the scene to provide initial estimates of the EOPs and IOPs of each image. The ellipses, representing the boundaries of the circular black and white targets, were detected from each image. A new method was then proposed to match the ellipses between different camera views, given the initial camera projective matrices provided by SfM. The center of each target, viewed by at least two images in the network, was then adjusted to correct for the eccentricity error using another newly developed method. Self-calibrating bundle adjustment was performed to re-estimate the EOPs and IOPs using the adjusted centers of each target. The EOPs and IOPs can be re-introduced to the previous steps for further iteration and refinement until the required precision for the adjusted centers is achieved (or until no more matches are found).

Four experiments were designed to assess the effectiveness of the proposed calibration methods using 4K video recordings captured via the iPhone 11, Huawei P30 and Samsung S10. The evaluation of the proposed center adjustment showed that the adjusted center provided around 15 times better 3D reconstructed center estimation accuracy, compared to when no center adjustment was performed. The results also revealed that as long as the camera matrices are available with a superior precision, five camera views can be sufficient to provide a sub-millimeter accuracy for the 3D reconstructed center of circular targets. It was also shown that the proposed method for estimating the eccentricity error in each image plane was comparable (and in some cases outperformed) the closed formulation, provided by (Luhmann, 2014). The effectiveness of the proposed method in correcting the eccentricity error provides opportunities to utilize larger targets.

The third experiment assessed the quality of the ellipse matching algorithm. It was demonstrated that the proposed method performed best when using both the robust triangulation to eliminate possible false matches (Type II errors), and the adjusted centers to increase the correct matches (enhancing Type I errors). The last experiment evaluated the accuracy of the estimated radius of a professionally installed mechanical mock pipework using both the proposed calibration parameters and the sequential SfM with in-situ (referred to here as COLMAP-radial). It was observed that for all mobile devices -i.e. iPhone 11, Huawei P30, and Samsung S10- the calibration parameters, estimated using our proposed method, provided a better accuracy, compared to COLMAP-radial (around 1.3 times better). The results of the experiments show that the proposed process for calibration is advantageous. Specifically, for the case of the mobile devices, the pre-calibration was found necessary to achieve better pipe radius estimation results.

## ACKNOWLEDGEMENTS

The authors wish to acknowledge the support provided by the MJS Mechanical Ltd. and Michael Baytalan for their professional installation of the mechanical pipes in the calibration laboratory for the purpose of this study, as well as supplementary discussions and feedback from GOLDBECK Ltd. This research project was funded by the Natural Sciences and Engineering Research Council (NSERC) of Canada (542980 - 19).